\documentclass[12pt]{article}
\usepackage{amsfonts}
\usepackage{a4wide}
\usepackage{amsmath,amscd,amsthm,a4,amssymb}

\begin{document}

\begin{center}
{\Large \textbf{Universal approximation theorem for neural networks with
inputs from a topological vector space}}

\

\textsc{Vugar E. Ismailov} \

\bigskip

Institute of Mathematics and Mechanics, Baku, Azerbaijan

Center for Mathematics and its Applications, Khazar University, Baku,
Azerbaijan

{e-mail:} {vugaris@mail.ru}
\end{center}

\smallskip

\textbf{Abstract.} We study feedforward neural networks with inputs from a
topological vector space (TVS-FNNs). Unlike traditional feedforward neural
networks, TVS-FNNs can process a broader range of inputs, including
sequences, matrices, functions and more. We prove a universal approximation
theorem for TVS-FNNs, which demonstrates their capacity to approximate any
continuous function defined on this expanded input space.

\bigskip



\

\begin{center}
{\large \textbf{1. Introduction}}
\end{center}

Neural networks have become a cornerstone of modern machine learning and
artificial intelligence, offering powerful tools for modeling and solving
complex problems. Among the various types of neural network architectures,
the multilayer feedforward perceptron (MLP) stands out as one of the most
widely used and fundamental models. The MLP is particularly valued for its
ability to approximate complex, nonlinear functions and perform tasks such
as classification, regression, and pattern recognition. This model is
structured with a finite number of sequential layers: the first layer is the
input layer, the last layer is the output layer, and the layers in between
are referred to as hidden layers. In the MLP, information flows from the
input layer through the hidden layers to the output layer. During this
process, each neuron in a layer receives inputs from neurons in the
preceding layer, applies a weight to these inputs, adds a bias, and then
passes the result through an activation function. The activation function
introduces non-linearity into the model, enabling it to learn and represent
complex patterns. The output of each neuron in one layer becomes the input
for the neurons in the subsequent layer. This sequential flow continues
until the final output is produced by the output layer.

The simplest case of an MLP is \textit{a single hidden layer neural network}%
. In this configuration, each output neuron computes a function of the form:
\begin{equation*}
\sum_{i=1}^{r}c_{i}\sigma (\mathbf{w}^{i}\cdot \mathbf{x}-\theta _{i}),\eqno%
(1.1)
\end{equation*}%
where $\mathbf{x}=(x_{1},...,x_{d})$ represents the input vector, $r$ is the
number of neurons in the hidden layer, $\mathbf{w}^{i}$ are \textit{weight
vectors} in $\mathbb{R}^{d}$, $\theta _{i}$ are \textit{thresholds}, $c_{i}$
are coefficients, and $\sigma $ is the \textit{activation function}, a real
univariate function. In this model, the activation function $\sigma $
introduces non-linearity, allowing the network to approximate more complex
functions by combining the outputs from the hidden layer neurons.

The theoretical foundation for neural networks is the \textit{universal
approximation property} (UAP), also known as the density property. This
property asserts that a neural network with a single hidden layer can
approximate any continuous function on a compact domain to any desired level
of accuracy. That is, the set $span\{\sigma (\mathbf{w\cdot x}-\theta
):\theta \in \mathbb{R},\mathbf{w}\in \mathbb{R}^{d}\}$, which consists of
functions of the form given in equation (1.1), is dense in $C(K)$ for every
compact set $K\subset \mathbb{R}^{d}$. In neural network theory, this result
is known as the \textit{universal approximation theorem} (UAT). There is a
substantial body of research addressing UAT for various activation functions
$\sigma $. Numerous results have explored how different choices of
activation functions affect the approximation capabilities of neural
networks. For example, see the papers \cite{Chen,Chui,Cot,Cyb,Fun,Hor,Ito}.
The most general result of this type was obtained by Leshno, Lin, Pinkus and
Schocken \cite{Leshno}. They proved that a continuous activation function $%
\sigma $ possesses the density property if and only if it is not a
polynomial. This result highlights the effectiveness of the single hidden
layer perceptron model for a wide range of activation functions $\sigma $.
It should be remarked that, in fact, the universal approximation theorem in
\cite{Leshno} was proved for a broader class of functions than just
continuous functions. This includes activation functions that may have
discontinuities on sets of Lebesgue measure zero. However, in this paper, we
restrict our focus to continuous activation functions. For a detailed,
step-by-step proof of this theorem, see also \cite{Pet,Pinkus}.

Previously, it was widely believed and emphasized in many studies that
achieving the universal approximation property requires large networks with
sufficiently many hidden neurons (see, e.g., \cite[Chapter 6.4.1]{Good}).
Note that in the aforementioned earlier papers, the number of hidden neurons
was considered unbounded. However, more recent studies \cite{GI2,GI3,Ism}
have shown that neural networks with practically computable activation
functions and a very small number of hidden neurons can approximate any
continuous function on any compact set to an arbitrary degree of accuracy.

Note that the inner product $\mathbf{w}^{i}\cdot \mathbf{x}$ in (1.1)
represents a linear continuous functional on $\mathbb{R}^{d}$. This is a
well known result in functional analysis that every linear continuous
functional $f$ on a Hilbert space $H$ can be expressed as $f(x)=\left\langle
a,x\right\rangle ,$ where $a$ is a unique element of $H$ and $\left\langle
\cdot ,\cdot \right\rangle $ denotes the inner product (see \cite[Theorem
13.32]{Roman}). Consequently, every linear functional on $\mathbb{R}^{d}$ is
of the form $\mathbf{w\cdot x}$, where $\mathbf{w}\in \mathbb{R}^{d}$ and $%
\mathbf{x}=(x_{1},...,x_{d})$ is the variable. This observation tells the
following generalization of single hidden layer networks from $\mathbb{R}%
^{d} $ to any topological vector space $X$. The architecture of such a
generalized feedforward neural network is as follows:

\begin{itemize}
\item \textbf{Input Layer:} This layer consists of an element $x\in X$. Note
that in the case where $X=\mathbb{R}^{d}$, $x$ is a $d$-dimensional vector.
Thus, only in this particular case, the layer has $d$ neurons, each
receiving an input signal $x_{1},...,x_{d}$.

\item \textbf{Hidden layer:} Each neuron in the hidden layer receives the
input $x$ from the input layer and applies a functional $f\in X^{\ast }$ to $%
x $ (where $X^{\ast }$ is the dual space of linear continuous functionals on
$X $). A shift $\theta $ and then a fixed activation function $\sigma :$ $%
\mathbb{R\rightarrow R}$ are applied to $f(x)$ and the resulting value $%
\sigma (f(x)-\theta )$ represents the output signal of the neuron.

\item \textbf{Output layer:} Each neuron in this layer receives weighted
signals from each neuron in the hidden layer, sums them, and produces the
final output value.
\end{itemize}

In the following discussion, the term \textquotedblleft feedforward neural
network with inputs from a topological vector space" will be abbreviated as
TVS-FNN. Note that in the architecture of TVS-FNNs, the element $x\in X$
carries all the information of the input layer. This structure allows the
network to handle a wide range of input types. In particular, $x$ may be an
infinite dimensional vector $(x_{1},x\,_{2},...)$, indicating an input with
infinitely many signals, or it may be even a function. Detailed examples
will be provided later in the form of corollaries from the main result (see
Section 3).

Thus a TVS-FNN with a single hidden layer evaluates a function of the form
\begin{equation*}
\sum_{i=1}^{r}c_{i}\sigma (f_{i}(x)-\theta _{i}),\eqno(1.2)
\end{equation*}%
where $x\in X$ is the input, $f_{i}\in X^{\ast },$ $c_{i},\theta _{i}\in
\mathbb{R}$ are the parameters of the network, and $\sigma :$ $\mathbb{%
R\rightarrow R}$ is a fixed activation function.

The aim of this paper is to show that for broad classes of topological
vector spaces $X$ and activation functions $\sigma $, neural networks of the
form presented in (1.2) can approximate any continous function on a compact
subset $K\subset X$ with arbitrary precision. In other words, the set
\begin{equation*}
\mathcal{N}_{1}(\sigma )=span\{\sigma (f(x)-\theta ):f\in X^{\ast },\theta
\in \mathbb{R}\}
\end{equation*}%
is dense in $C(K)$ for every compact set $K\subset X$.

Let us remark that TVS-FNNs with more than one hidden layer are defined
recursively, similar to traditional networks. Thus, TVS-FNNs with $l$ hidden
layers can be represented as elements of the space
\begin{equation*}
\mathcal{N}_{l}(\sigma )=span\{\sigma (G(x)-\theta ):G\in \mathcal{N}%
_{l-1}(\sigma ),\theta \in \mathbb{R}\}.
\end{equation*}%
However, this paper does not address the approximation capabilities of deep
TVS-FNNs.

It should be remarked that UAP of neural networks acting between Banach
spaces has been investigated in several studies. For example, in \cite{Sun},
the fundamentality of ridge functions was established in a Banach space and
then applied to shallow networks with a sigmoidal activation function (see
also \cite{Light}). In \cite{Chen2}, the authors showed that any continuous
nonlinear function mapping a compact set $V$ in a Banach space of continuous
functions $C(K_{1})$ into $C(K_{2})$ can be approximated arbitrarily well by
shallow feedforward neural networks. Here $K_{1}$ and $K_{2}$ are two
compact sets in an abstract Banach space $X$ and the Euclidean space $%
\mathbb{R}^{d}$, respectively. In \cite{Lu}, this construction was extended
to deep neural networks and named DeepONet. In \cite{Lant}, the authors
analyzed DeepONet within the framework of an encoder-decoder type of network
and examined its approximation properties for the case where the input space
is a Hilbert space. In \cite{Kor}, quantitative estimates (i.e., convergence
rates) for the approximation of nonlinear operators using single-hidden
layer networks acting between infinite-dimensional Banach spaces were
obtained, extending some existing results from the finite-dimensional case.

Recent studies have established UAT for certain hypercomplex-valued neural
networks such as complex-, quaternion-, tessarine-, and Clifford-valued
neural networks, as well as more general vector-valued neural networks
(V-nets) defined on and with values from a finite-dimensional algebra (see
\cite{Valle} and references therein). We hope that the findings of this
paper will encourage further research on TVS-FNNs with outputs from these
and other general spaces.

\bigskip

\begin{center}
{\large \textbf{2. Main result}}
\end{center}

In this section, we show that shallow TVS-FNNs, when using any
non-polynomial continuous activation function, possess the universal
approximation property. This result generalizes the well-known theorem by
Leshno, Lin, Pinkus, and Schocken \cite{Leshno}. In the proof, we utilize
their result in the one-dimensional case. For completeness, we also provide
a short proof for this special case.

In the following theorem, we consider a topological vector space (TVS)\ with
the \textit{Hahn-Banach Extension Property} (HBEP). A TVS $X$ is said to
have the HBEP if every continuous linear functional defined on a closed
subspace of $X$ can be extended to the whole space while remaining
continuous. According to the Hahn-Banach theorem, every (locally) convex TVS
has the HBEP (see \cite[Theorem 3.6]{Rudin}). However, there also exist
nonconvex TVSs that possess the HBEP (see \cite{Greg}).

\bigskip

\textbf{Theorem 2.1.} \textit{Assume $X$ is a TVS with the HBEP, $\Theta $
is any open interval in $\mathbb{R}$ and $\sigma$ is a continuous univariate
function that is not a polynomial on $\Theta $. Then for any $\varepsilon >0$%
, any compact set $K\subset X$ and any function $g\in C(K)$ there exist $%
r\in \mathbb{N}$, $f_{i}\in X^{\ast }$, $c_{i}\in \mathbb{R}$, $\theta
_{i}\in \Theta $, $i=1,...,r$, such that}
\begin{equation*}
\max_{x\in K}\left\vert g(x)-\sum_{i=1}^{r}c_{i}\sigma (f_{i}(x)-\theta
_{i})\right\vert <\varepsilon .
\end{equation*}

\bigskip

\textbf{Proof.} Assume $\sigma $ is not a polynomial on $\Theta $. The proof
is carried out in four steps.

\textit{Step 1: The case when }$X=\mathbb{R}$\textit{\ and }$\sigma \in
C^{\infty }(\mathbb{R})$. We first prove that if $\sigma \in C^{\infty }(%
\mathbb{R})$, then the set
\begin{equation*}
\mathcal{M}(\sigma )=span\{\sigma (w\cdot x-\theta ):w\in \mathbb{R},\theta
\in \Theta \}
\end{equation*}%
is dense in $C(K)$ for every compact set $K\subset \mathbb{R}$. To show
this, note that

\begin{equation*}
\left[ \sigma ((w+h)x-\theta )-\sigma (wx-\theta )\right] /h\in \mathcal{M}%
(\sigma )
\end{equation*}%
for every $w\in \mathbb{R}$, $\theta \in \Theta $ and $h\neq 0$. Since $%
\sigma \in C^{\infty }(\mathbb{R})$, it follows that $\frac{d}{dw}\sigma
(wx-\theta )\in \overline{\mathcal{M}(\sigma )}$. By the same argument $%
\frac{d^{k}}{dw^{k}}\sigma (wx-\theta )\in \overline{\mathcal{M}(\sigma )}$
for all $k\in \mathbb{N}$ and all $w\in \mathbb{R}$, $\theta \in \Theta $.
Obviously,

\begin{equation*}
\frac{d^{k}}{dw^{k}}\sigma (wx-\theta )=x^{k}\sigma ^{(k)}(wx-\theta )
\end{equation*}%
and since $\sigma $ is not a polynomial, there exists $-\theta _{k}\in
\Theta $ such that $\sigma ^{(k)}(-\theta _{k})\neq 0$. Therefore,

\begin{equation*}
x^{k}\sigma ^{(k)}(-\theta _{k})=x^{k}\sigma ^{(k)}(wx-\theta )|_{w=0,\theta
=\theta _{k}}\in \overline{\mathcal{M}(\sigma )}.
\end{equation*}%
We see that $\overline{\mathcal{M}(\sigma )}$ contains all monomials and
thus all algebraic polynomials. By the Weierstrass theorem this implies that
$\mathcal{M}(\sigma )$ is dense in $C(K)$ for every compact set $K\subset
\mathbb{R}$.

\bigskip

\textit{Step 2: The case when }$X=\mathbb{R}$\textit{\ and }$\sigma \in C(%
\mathbb{R})$. Let us now prove that if $\sigma \in C(\mathbb{R})$, then the
set $\mathcal{M}(\sigma )$ is dense in $C(K)$ for any compact set $K\subset
\mathbb{R}$. We prove this assertion by contradiction. Assume, on the
contrary, that $\mathcal{M}(\sigma )$ is not dense. Consider the convolution
of $\sigma $ with a function $\phi \in C_{0}^{\infty }(\mathbb{R})$, that
is, an infinitely differentiable function with compact support
\begin{equation*}
(\sigma \ast \varphi )(t)=\int\limits_{-\infty }^{+\infty }\sigma (t-y)\phi
(y)dy.
\end{equation*}

Since $\sigma $ and $\varphi $ are continuous functions and $\varphi $ has
compact support the above integral converges for all $t$. Taking Riemann
sums, it is easy to understand that $(\sigma \ast \varphi )(t)$ is contained
in $\overline{\mathcal{M}(\sigma )}$. By the same way, we obtain that $%
(\sigma \ast \varphi )(wt-\theta )$ is contained in $\overline{\mathcal{M}%
(\sigma )}$. Thus, $\overline{\mathcal{M}(\sigma \ast \varphi )}\subset
\overline{\mathcal{M}(\sigma )}$. Since $\sigma \ast \varphi \in C^{\infty }(%
\mathbb{R}),$ we have from Step 1 that $t^{k}(\sigma \ast \varphi
)^{(k)}(-\theta )\in \overline{\mathcal{M}(\sigma \ast \varphi )}$ for all $%
k\in \mathbb{N}$ and all $\theta \in \Theta $. Since $\mathcal{M}(\sigma )$
is not dense, the monomial $t^{k}$ is not in $\overline{\mathcal{M}(\sigma
\ast \varphi )}$ for some $k$. Therefore, $t^{k}$ is not in $\overline{%
\mathcal{M}(\sigma \ast \varphi )}$ for each $\varphi \in C_{0}^{\infty }(%
\mathbb{R})$. This means that $(\sigma \ast \varphi )^{(k)}(-\theta )=0$ for
all $\theta \in \Theta $. Hence for any $\varphi \in C_{0}^{\infty }(\mathbb{%
R})$ the convolution $\sigma \ast \varphi $ is a polynomial of degree at
most $k-1$ on $\Theta $.

It is well known that there exists a sequence $\{\varphi
_{n}\}_{n=1}^{\infty }\subset C_{0}^{\infty }(\mathbb{R})$ for which $\sigma
\ast \varphi _{n}$ converges uniformly to $\sigma $ on any compact set in $%
\mathbb{R}$. Since all $\sigma \ast \varphi _{n}$ are polynomials of degree
at most $k-1$, the limit must be also a polynomial of degree at most $k-1$.
Thus we obtain that $\sigma $ is a polynomial on $\Theta $. This contradicts
our assumption.

\bigskip

\textit{Step 3: Exponential functions.} Let us show that the set

\begin{equation*}
E=span\{e^{r(x)}:r\in X^{\ast }\}.
\end{equation*}%
is dense in $C(K)$ for every compact set $K\subset X$.

It is not difficult to see that $E$ is a subalgebra of $C(X)$. Indeed, for
any $r_{1},r_{2}\in X^{\ast }$

\begin{equation*}
e^{r_{1}(x)}e^{r_{2}(x)}=e^{r_{1}(x)+r_{2}(x)}\in E\text{,}
\end{equation*}%
since $r_{1}+r_{2}\in X^{\ast }$. Therefore, the linear space $E$ is closed
under multiplication, indicating that $E$ is an algebra.

Moreover, if $r$ is the zero functional, then $e^{r(x)}=1$, showing that $E$
contains all constant functions. Since $X$ has the HBEP, for any distinct
points $x_{1},x_{2}\in X$ there exists a functional $r\in X^{\ast }$ such
that $r(x_{1})\neq r(x_{2})$. Hence, the algebra $E$ separates points in $X$.

By the Stone-Weierstrass theorem \cite{Stone}, for any compact $K\subset X$,
the algebra $E$ restricted to $K$ is dense in $C(K)$.

\bigskip

\textit{Step 4: The general case.} According to Step 3 for any compact $%
K\subset X$, any $g\in C(K)$ and any $\varepsilon >0$, there exist finitely
many functionals $r_{i}\in X^{\ast }$ and numbers $a_{i}\in \mathbb{R}$ such
that

\begin{equation*}
\left\vert g(x)-\sum_{i=1}^{n}a_{i}e^{r_{i}(x)}\right\vert <\varepsilon /2%
\eqno(2.1)
\end{equation*}%
for all $x\in K$. Since $r_{i}$ are continuous, the images $r_{i}(K)$ are
compact sets in $\mathbb{R}$. Put $R=\cup _{i=1}^{n}r_{i}(K)$. Note that $R$
is also compact. By Step 2, each univariate function $a_{i}e^{t}$, $t$ $\in
R $, can be approximated by single hidden layer networks with the activation
$\sigma $. Thus, there exist coefficients $c_{ij},w_{ij}\in \mathbb{R}$, $%
\theta _{ij}\in \Theta $, $1\leq i\leq n$, $1\leq j\leq k$, such that

\begin{equation*}
\left\vert a_{i}e^{t}-\sum_{j=1}^{k}c_{ij}\sigma (w_{ij}t-\theta
_{ij})\right\vert <\varepsilon /2n
\end{equation*}%
for all $t\in R$. Hence,

\begin{equation*}
\left\vert a_{i}e^{r_{i}(x)}-\sum_{j=1}^{k}c_{ij}\sigma
(w_{ij}r_{i}(x)-\theta _{ij})\right\vert <\varepsilon /2n\eqno(2.2)
\end{equation*}%
for each $i=1,...,n$, and all $x\in K$. It follows from (2.1) and (2.2) that

\begin{equation*}
\left\vert g(x)-\sum_{i=1}^{n}\sum_{j=1}^{k}c_{ij}\sigma
(w_{ij}r_{i}(x)-\theta _{ij})\right\vert <\varepsilon
\end{equation*}%
for any $x\in K$. This completes the proof of Theorem 2.1.

\bigskip

\textbf{Remark.} Let $B$ be a bounded set in $X$ such that no nontrivial
linear functional vanishes on $B$. In Theorem 1, $X^{\ast }$ can be replaced
by a subset $S=\{f\in X^{\ast }:\sup_{x\in B}\left\vert f(x)\right\vert =1\}$%
. To prove this, instead of using $r_{i}(x)$ in (2.1), let us use $\alpha
_{i}v_{i}(x)$, where $\alpha _{i}=\sup_{x\in B}\left\vert
r_{i}(x)\right\vert $ and $v_{i}(x)=\frac{1}{\alpha _{i}}r_{i}(x)$. We then
obtain

\begin{equation*}
\left\vert g(x)-\sum_{i=1}^{n}u_{i}(v_{i}(x))\right\vert <\varepsilon /2,
\end{equation*}%
where $u_{i}(t)=a_{i}e^{\alpha _{i}t}$ are continuous univariate functions.

Since $v_{i}$ are continuous, the images $v_{i}(K)$ are compact sets in $%
\mathbb{R}$. Put $V=\cup _{i=1}^{n}v_{i}(K)$. By Step 2, each univariate
function $u_{i}(t)$, $t$ $\in V$, can be approximated by single hidden layer
networks with the activation $\sigma $. The remainder of the proof follows
similarly to Step 4 above. This remark implies, for instance, that in the
classical UAT all weights can be taken from a sphere (in particular, $%
B\subset X$ can be a sphere).

\bigskip

\begin{center}
{\large \textbf{3. Some corollaries}}
\end{center}

We present several corollaries of Theorem 2.1, considering various
topological vector spaces $X$, which are also Banach spaces. These
corollaries rely on the structure of the continuous dual $X^{\ast }$ of the
considered $X.$ The form of functionals in such $X^{\ast }$ is well-known
and extensively studied in Linear Algebra and Functional Analysis textbooks.
Before stating the results, we note that in $C(X)$, we will use the topology
of uniform convergence on compact sets. This topology is induced by the
seminorms

\begin{equation*}
\left\Vert g\right\Vert _{K}=\max_{x\in K}\left\vert g(x)\right\vert ,
\end{equation*}%
where $K$ are compact sets in $X$. A subbasis at the origin for this
topology is given by the sets%
\begin{equation*}
U(K,r)=\left\{ g\in C(X):\left\Vert g\right\Vert _{K}<r\right\} ,
\end{equation*}%
where $K\subset X$ is compact and $r>0$. A sequence (or net) $\{g_{n}\}$ in
this topology converges to $g$ iff $\left\Vert g_{n}-g\right\Vert
_{K}\rightarrow 0$ for every compact set $K\subset X$. Thus, in what
follows, when we say that $B$ is dense in $C(X)$, we will mean that $B$ is
dense with respect to the aforementioned topology of uniform convergence on
compact sets.

The following corollaries are valid. For simplicity, we will assume that the
set of thresholds $\Theta =\mathbb{R}$.

\bigskip

\textbf{Corollary 3.1.} \textit{Let $M_{n\times m}(\mathbb{K})$ be the
vector space of $n\times m$ matrices over a field $\mathbb{K}$ and $\sigma $
be any continuous non-polynomial activation function. Then the set of
TVS-FNNs}
\begin{equation*}
\mathcal{N}(\sigma ,M_{n\times m}(\mathbb{K}))=span\{\sigma
(trace(W^{T}X)-\theta ):W\in M_{n\times m}(\mathbb{K}),\theta \in \mathbb{R}%
\},
\end{equation*}%
\textit{with inputs $X\in M_{n\times m}(\mathbb{K})$, is dense in $%
C(M_{n\times m}(\mathbb{K}))$.}

\bigskip

\textbf{Corollary 3.2}. \textit{Let $1\leq p<\infty $ be a real number and $%
l_{p}$ be the Banach space of all sequences $\mathbf{x}=(x_{n})$ for which}
\begin{equation*}
\left\Vert \mathbf{x}\right\Vert _{p}=\left( \sum_{n=1}^{\infty }\left\vert
x_{n}\right\vert ^{p}\right) ^{\frac{1}{p}}<\infty .
\end{equation*}%
\textit{Assume $\sigma $ is any continuous non-polynomial activation
function and let $q=\frac{p}{p-1}$. Then the set of TVS-FNNs}
\begin{equation*}
\mathcal{N}(\sigma ,l_{p})=span\left\{ \sigma (\sum_{n=1}^{\infty
}a_{n}x_{n}-\theta ):(a_{n})\in l_{q},\theta \in \mathbb{R}\right\} ,
\end{equation*}%
\textit{with inputs $(x_{n})\in l_{p}$, is dense in $C(l_{p})$.}

\bigskip

\textbf{Corollary 3.3.} \textit{Let $c_{0}$ denote the space of all
sequences $(x_{n})$ that converge to zero and let $\sigma $ be any
continuous non-polynomial activation function. Then the set of TVS-FNNs}
\begin{equation*}
\mathcal{N}(\sigma ,c_{0})=span\left\{ \sigma (\sum_{n=1}^{\infty
}a_{n}x_{n}-\theta ):(a_{n})\in l_{1},\theta \in \mathbb{R}\right\} ,
\end{equation*}%
\textit{with inputs $(x_{n})\in c_{0}$, is dense in $C(c_{0})$.}

\bigskip

\textbf{Corollary 3.4.} \textit{Let $1\leq p<\infty $ be a real number and $%
L_{p}(X,\mu )$ be the space of all $\mu $-measurable functions $%
f:X\rightarrow \mathbb{R}$ such that the $p$-norm is finite, i.e.,}
\begin{equation*}
\left\Vert f\right\Vert _{p}=\left( \int_{X}\left\vert f(x)\right\vert
^{p}d\mu \right) ^{\frac{1}{p}}<\infty \text{.}
\end{equation*}%
\textit{Assume $\sigma $ is any continuous non-polynomial activation
function and let $q=\frac{p}{p-1}$. Then the set of TVS-FNNs}
\begin{equation*}
\mathcal{N}(\sigma ,L_{p}(X,\mu ))=span\left\{ \sigma \left(
\int_{X}f(x)g(x)d\mu -\theta \right) :g(x)\in L_{q}(X,\mu ),\theta \in
\mathbb{R}\right\} ,
\end{equation*}%
\textit{with inputs $f\in L_{p}(X,\mu )$, is dense in $C(L_{p}(X,\mu ))$.}

\bigskip

\textbf{Corollary 3.5.} \textit{Let $X$ be a compact Hausdorff space, $C(X)$
be the Banach space of continuous functions on $X$ and $\mathcal{B}(X)$ be
the class of regular real-valued measures of finite total variation defined
on Borel subsets of $X$. Assume $\sigma $ is any continuous non-polynomial
activation function. Then the set of TVS-FNNs}

\begin{equation*}
\mathcal{N}(\sigma ,C(X))=span\left\{ \sigma \left( \int_{X}f(x)d\mu -\theta
\right) :\mu \in \mathcal{B}(X),\theta \in \mathbb{R}\right\} ,
\end{equation*}%
\textit{with inputs $f\in C(X)$, is dense in $C(C(X))$.}

\bigskip

It should be remarked that in the above corollaries and in Theorem 2.1, we
can use any activation function $\sigma :\mathbb{R}\rightarrow \mathbb{R}$
(whether continuous or discontinuous) with the property that the $%
span\{\sigma (wx-\theta ):w\in \mathbb{R},\theta \in \mathbb{R}\}$ is dense
in every $C[a,b]$. Such functions are called Tauber-Wiener (TW) functions
(see \cite{Chen2}). It follows from the main result of \cite{Leshno} that a
continuous nonpolynomial function is a TW function.

TVS-FNNs with inputs from $C(X)$ was also discussed in \cite{Chen2}. Note
that Corollary 3.5 is stronger than the result in \cite[Theorem 4]{Chen2},
which included the expression $\sigma \left( \sum_{j}\xi _{j}f(x_{j})-\theta
\right) $, with $\xi _{j}\in \mathbb{R}$, $x_{j}\in X$, in place of $\sigma
\left( \int_{X}f(x)d\mu -\theta \right) $.


\bigskip

\begin{thebibliography}{99}
\bibitem{Chen} T. Chen and H. Chen, Approximation of continuous functionals
by neural networks with application to dynamic systems, \textit{IEEE Trans.
Neural Networks} \textbf{4} (1993), 910-918.

\bibitem{Chen2} T. Chen and H. Chen, Universal approximation to nonlinear
operators by neural networks with arbitrary activation functions and its
application to dynamical systems, \textit{IEEE Trans. Neural Netw.} \textbf{6%
} (1995), no. 4, 911-917.

\bibitem{Chui} C. K. Chui and X. Li, Approximation by ridge functions and
neural networks with one hidden layer, \textit{J. Approx. Theory} \textbf{70}
(1992), 131-141.

\bibitem{Cot} N. E. Cotter, The Stone-Weierstrass theorem and its
application to neural networks, \textit{IEEE Trans. Neural Netw.} \textbf{1}
(1990), 290-295.

\bibitem{Cyb} G. Cybenko, Approximation by superpositions of a sigmoidal
function, \textit{Math. Control, Signals, and Systems} \textbf{2} (1989),
303-314.

\bibitem{Fun} K. Funahashi, On the approximate realization of continuous
mapping by neural networks, \textit{Neural Netw.} \textbf{2} (1989), 183-192.

\bibitem{Good} I. Goodfellow, Y. Bengio and A. Courville, \textit{Deep
Learning}, MIT Press, Cambridge, MA, 2016.

\bibitem{Greg} D. A. Gregory and J. H. Shapiro, Nonconvex linear topologies
with the Hahn Banach extension property, \textit{Proc. Amer. Math. Soc.}
\textbf{25} (1970), 902-905.

\bibitem{GI2} N. J. Guliyev and V. E. Ismailov, On the approximation by
single hidden layer feedforward neural networks with fixed weights, \textit{%
Neural Netw.} \textbf{98} (2018), 296-304.

\bibitem{GI3} N. J. Guliyev and V. E. Ismailov, Approximation capability of
two hidden layer feedforward neural networks with fixed weights, \textit{%
Neurocomputing} \textbf{316} (2018), 262-269.

\bibitem{Hor} K. Hornik, Approximation capabilities of multilayer
feedforward networks, \textit{Neural Networks} \textbf{4} (1991), 251-257.

\bibitem{Ism} V. E. Ismailov, \textit{Ridge Functions and Applications in
Neural Networks}, Mathematical Surveys and Monographs, 263. American
Mathematical Society, 2021.

\bibitem{Ito} Y. Ito, Approximation of continuous functions on $\mathbb{R}%
^{d}$ by linear combinations of shifted rotations of a sigmoid function with
and without scaling, \textit{Neural Netw.} \textbf{5} (1992), 105-115.

\bibitem{Kor} Y. Korolev, Two-layer neural networks with values in a Banach
space, \textit{SIAM J. Math. Anal.} \textbf{54} (2022), no. 6, 6358-6389.

\bibitem{Lant} S. Lanthaler, S. Mishra and G. E. Karniadakis, Error
estimates for DeepONets: a deep learning framework in infinite dimensions,
\textit{Trans. Math. Appl.} \textbf{6} (2022), no. 1, tnac001, 141 pp.

\bibitem{Leshno} M. Leshno, V. Ya. Lin, A. Pinkus and S. Schocken,
Multilayer feedforward networks with a nonpolynomial activation function can
approximate any function, \textit{Neural Netw.} \textbf{6} (1993), 861-867.

\bibitem{Light} W. Light, Ridge functions, sigmoidal functions and neural
networks, Approximation theory VII (Austin, TX, 1992), 163-206, Academic
Press, Boston, MA, 1993.

\bibitem{Lu} L. Lu, P. Jin, G. Pang, Z. Zhang and G. E. Karniadakis,
Learning nonlinear operators via DeepONet based on the universal
approximation theorem of operators, \textit{Nat. Mach. Intell.} \textbf{3}
(2021), no. 3, 218-229.

\bibitem{Pet} P. Petersen and J. Zech, Mathematical theory of deep learning,
arXiv preprint, arXiv:2407.18384 [cs.LG], 2024.

\bibitem{Pinkus} A. Pinkus, Approximation theory of the MLP model in neural
networks, \textit{Acta Numerica} \textbf{8} (1999), 143-195.

\bibitem{Roman} S. Roman, \textit{Advanced linear algebra}, Third edition.
Graduate Texts in Mathematics, 135. Springer, New York, 2008.

\bibitem{Rudin} W. Rudin, \textit{Functional analysis}, Second edition.
International Series in Pure and Applied Mathematics. McGraw-Hill, Inc., New
York, 1991, 424 pp.

\bibitem{Stone} M. H. Stone, The generalized Weierstrass approximation
theorem, \textit{Math. Mag.} \textbf{21} (1948), 167-184, 237-254.

\bibitem{Sun} X. Sun, E. W. Cheney, The fundamentality of sets of ridge
functions, \textit{Aequationes Math.} \textbf{44} (1992), no. 2-3, 226-235.

\bibitem{Valle} M. E. Valle, W. L. Vital and G. Vieira, Universal
approximation theorem for vector- and hypercomplex-valued neural networks,
\textit{Neural Netw.} \textbf{180} (2024), 106632.
\end{thebibliography}
\end{document}